\documentclass[5p]{elsarticle}

\usepackage{lineno,hyperref}
\usepackage{soul}
\soulregister\cite7 % 针对\cite命令
\soulregister\citep7 % 针对\citep命令
\soulregister\citet7 % 针对\citet命令
\soulregister\ref7 % 针对\ref命令
\soulregister\pageref7 % 针对\pageref命令
\usepackage{mathrsfs}
\usepackage[noend]{algpseudocode}

\usepackage{algorithmicx,algorithm}
\usepackage{bbm}
\usepackage{graphicx}
\usepackage{amsmath}
\usepackage{amssymb}
\usepackage{multirow}
\usepackage{helvet}
\usepackage{courier}
\usepackage{float}
\usepackage{graphicx}

\usepackage{amsfonts}
\usepackage{textcomp}
\usepackage{hyperref}

\usepackage{subfigure}
\usepackage{times}
\usepackage{epsfig}
\usepackage[usenames,dvipsnames]{color}

\usepackage{amsmath,amsfonts}

\usepackage{array}
\usepackage{textcomp}
\usepackage{stfloats}
\usepackage{url}
\usepackage{verbatim}
\usepackage{arydshln}
\usepackage{bbding}
\usepackage{wrapfig}
\usepackage{caption}
\usepackage{booktabs}
\usepackage{colortbl}
\usepackage{newfloat}
\usepackage{listings}
\usepackage{bm}
\usepackage{pifont}

\usepackage{tabularx}
\usepackage{booktabs}  % 美化表格
\usepackage{graphicx}  % 允许插入图片
\usepackage{adjustbox} % 让表格适应页面宽度
\definecolor{ceiling}{RGB}{210,42,33}
\definecolor{floor}{RGB}{40,159,7}
\definecolor{wall}{RGB}{162,215,224}
\definecolor{windows}{RGB}{113,158,201}
\definecolor{chair}{RGB}{202,205,76}
\definecolor{bed}{RGB}{224,190,159}
\definecolor{sofa}{RGB}{153,98,192}
\definecolor{table}{RGB}{23,124,181}
\definecolor{tvs}{RGB}{160,187,34}
\definecolor{furniture}{RGB}{225,126,18}
\definecolor{objects}{RGB}{199,174,221}

\definecolor{road}{RGB}{255,1,252}
\definecolor{sidewalk}{RGB}{76,0,74}
\definecolor{parking}{RGB}{255,149,255}
\definecolor{other_grnd}{RGB}{178,4,75}
\definecolor{building}{RGB}{255,198,0}
\definecolor{car}{RGB}{98,152,240}
\definecolor{truck}{RGB}{82,28,183}
\definecolor{bicycle}{RGB}{101,229,248}
\definecolor{motorcycle}{RGB}{35,57,148}
\definecolor{other_veh}{RGB}{100,82,245}
\definecolor{vegetation}{RGB}{2,176,0}
\definecolor{trunk}{RGB}{131,62,6}
\definecolor{terrain}{RGB}{153,238,85}
\definecolor{person}{RGB}{255,26,29}
\definecolor{bicyclist}{RGB}{255,40,202}
\definecolor{motorcycl}{RGB}{150,29,99}
\definecolor{fence}{RGB}{148,125,42}
\definecolor{pole}{RGB}{255,239,143}
\definecolor{traf_sign}{RGB}{248,2,3}

\makeatletter
\renewcommand{\@makefntext}[1]{#1}
\makeatother

\journal{Information Fusion}

\begin{document}

\begin{sloppypar}
\begin{frontmatter}

\title{HMR-1: Hierarchical Massage Robot with Vision-Language-Model for Embodied Healthcare}

\author{Rongtao Xu\textsuperscript{b,*}, 
        Mingming Yu\textsuperscript{c,*}, 
        Xiaofeng Han\textsuperscript{a,*}, 
        Yu Zhang\textsuperscript{a},
        Kaiyi Hu\textsuperscript{d},
        Zhe Feng\textsuperscript{a},
        Zenghuang Fu\textsuperscript{a},
        Changwei Wang\textsuperscript{e}, 
        Weiliang Meng\textsuperscript{a,\dag}, 
        Xiaopeng Zhang\textsuperscript{a}}

\address{%
\textsuperscript{a}The State Key Laboratory of Multimodal Artificial Intelligence Systems, Institute of Automation, Chinese Academy of Sciences, China\\[1ex]
\textsuperscript{b}Spatiotemporal AI, China\\[1ex]
\textsuperscript{c}Hangzhou International Innovation Institute, Beihang University, China\\[1ex]
\textsuperscript{d}Georgia Institute of Technology, China\\[1ex]
\textsuperscript{e}Key Laboratory of Computing Power Network and Information Security, Ministry of Education; Shandong Computer Science Center, Qilu University of Technology (Shandong Academy of Sciences), China}

\fntext[myfootnote]{\textsuperscript{*}Equal contribution.\\
\textsuperscript{\dag}Corresponding authors.}

\date{} % 这里使用空参数以去除日期

\begin{abstract}
The rapid advancement of Embodied Intelligence has opened transformative opportunities in healthcare, particularly in physical therapy and rehabilitation. However, critical challenges remain in developing robust embodied healthcare solutions, such as the lack of standardized evaluation benchmarks and the scarcity of open-source multimodal acupoint massage datasets. To address these gaps, we construct MedMassage-12K - a multimodal dataset containing 12,190 images with 174,177 QA pairs, covering diverse lighting conditions and backgrounds. Furthermore, we propose a hierarchical embodied massage framework, which includes a high-level acupoint grounding module and a low-level control module. The high-level acupoint grounding module uses multimodal large language models to understand human language and identify acupoint locations, while the low-level control module provides the planned trajectory. Based on this, we evaluate existing MLLMs and establish a benchmark for embodied massage tasks. Additionally, we fine-tune the Qwen-VL model, demonstrating the framework's effectiveness. Physical experiments further confirm the practical applicability of the framework.Our dataset and code are publicly available at \href{https://github.com/Xiaofeng-Han-Res/HMR-1}{https://github.com/Xiaofeng-Han-Res/HMR-1}.

\end{abstract}
\begin{keyword}
Embodied Intelligence, multimodal large language models, acupoint massage, healthcare robotics.
\end{keyword}

\end{frontmatter}
\section{Introduction}
\label{sec:intro}

Recent advancements in Multimodal Large Language Models (MLLMs)\cite{han2025multimodal,touvron2023llama,floridi2020gpt3, hurst2024gpt,jin2025efficient,yang2025thinking} and Embodied AI\cite{liu2025robomamba, black2410pi0,hou2025dita,xu20253d,Rdt-1b, zhang2024navid,lin2025evolvenav,zhang2025activevln,yan2024instrugen,lin2023advances,Rekep} have demonstrated remarkable capabilities in complex reasoning, human intention understanding, and robotic manipulation\cite{xu2025a0,zhang2025robridge,ma2025phyblock}. Models such as LLaVA \cite{liu2023visual}, Qwen-VL \cite{wang2024qwen2vl}, and ManipLLM \cite{li2024manipllm} have shown significant progress in integrating vision, language, and action, while robotics frameworks like RT-1 \cite{brohan2022rt1}, RT-2 \cite{brohan2023rt2}, and OpenVLA \cite{kim2024openvla} leverage large-scale pretrained models to directly predict robot actions. These breakthroughs suggest that embodied intelligence could transform domains requiring precise human-machine interaction.

Healthcare is one of the most promising application areas, particularly in physical therapy and rehabilitation. Intelligent systems can assist or even replace human therapists in repetitive and precise procedures, such as acupoint massage. However, developing robust embodied healthcare solutions remains challenging\cite{grammatikopoulou2021cadis,xie2024medtrinity, singhal2023large}. Current intelligent systems achieve strong results in passive tasks such as medical QA or imaging analysis, but struggle in treatment scenarios that demand active physical interaction and fine-grained grounding. Existing frameworks often rely on pre-programmed trajectories and fail to translate natural language into precise embodied actions\cite{cleary2019manipulation,chapman2018evidence,longhini2025unfolding,ye2025deepseek}.

% 医疗健康是其中最具潜力的应用方向之一，尤其是在物理治疗与康复方面。智能系统可以在重复且需要精确操作的程序中辅助甚至替代人类治疗师，例如穴位按摩。然而，开发稳健的具身医疗解决方案仍面临挑战。目前的智能系统在医学问答或影像分析等被动任务上表现优异，但在需要主动物理交互和细粒度落地的治疗场景中表现不足。现有框架通常依赖预设轨迹，难以将自然语言转化为精确的具身动作。

A straightforward alternative would be to adopt traditional object detection frameworks (e.g., Faster R-CNN \cite{ren2015faster}, YOLO\cite{yolov11}, RT-DETR\cite{RT-DETR}) for acupoint localization. Yet, these detectors are designed for static category prediction and bounding-box regression, making them inadequate for understanding complex human instructions. In acupoint massage, commands often involve both semantic interpretation and action grounding (e.g., ``locate the Zusanli acupoint and apply moderate pressure’’). Traditional detectors cannot align linguistic intent with visual perception, thus limiting their applicability. We aim to overcome this gap by leveraging MLLMs, which inherently combine language comprehension, visual grounding, and reasoning, enabling instruction-driven embodied massage.

% 一种直接的替代方案是采用传统的目标检测框架（如 Faster R-CNN、YOLO、RT-DETR）进行穴位定位。然而，这些检测器主要用于静态类别预测与边界框回归，难以理解复杂的人类指令。在穴位按摩场景中，指令往往涉及语义理解与动作落地（例如：“定位足三里穴并施加适度压力”）。传统检测器无法将语言意图与视觉感知对齐，从而限制了其适用性。为此，我们旨在利用多模态大语言模型（MLLMs）来弥补这一差距，其天然具备语言理解、视觉定位与推理能力，从而支持基于指令驱动的具身按摩。

\begin{figure}[t!]
  \centering
  \includegraphics[width=0.5\textwidth]{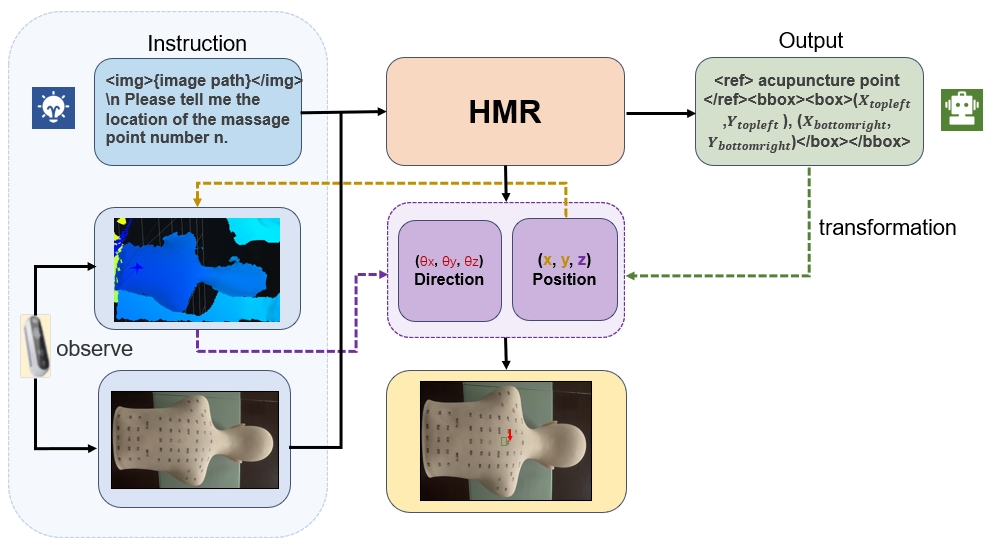}
  \caption{The proposed HMR framework. Given a textual instruction and an RGB-D observation, the model predicts the 6-DOF end-effector pose for robotic massage.}
  \vspace{-0.53cm}
\end{figure}

To address these challenges, we introduce \textbf{MedMassage-12K}, a multimodal dataset containing 12,190 images and 174,177 QA pairs under diverse lighting and background conditions, specifically designed for acupoint massage tasks. Building on this dataset, we propose a hierarchical embodied massage framework with two core components: (i) a high-level acupoint grounding module that employs MLLMs to understand natural language and localize acupoints, and (ii) a low-level control module that generates precise trajectories for robotic execution.\textbf{Our primary contributions are as follows:}
\begin{figure*}[t]
  \centering
  \includegraphics[width=1\textwidth]{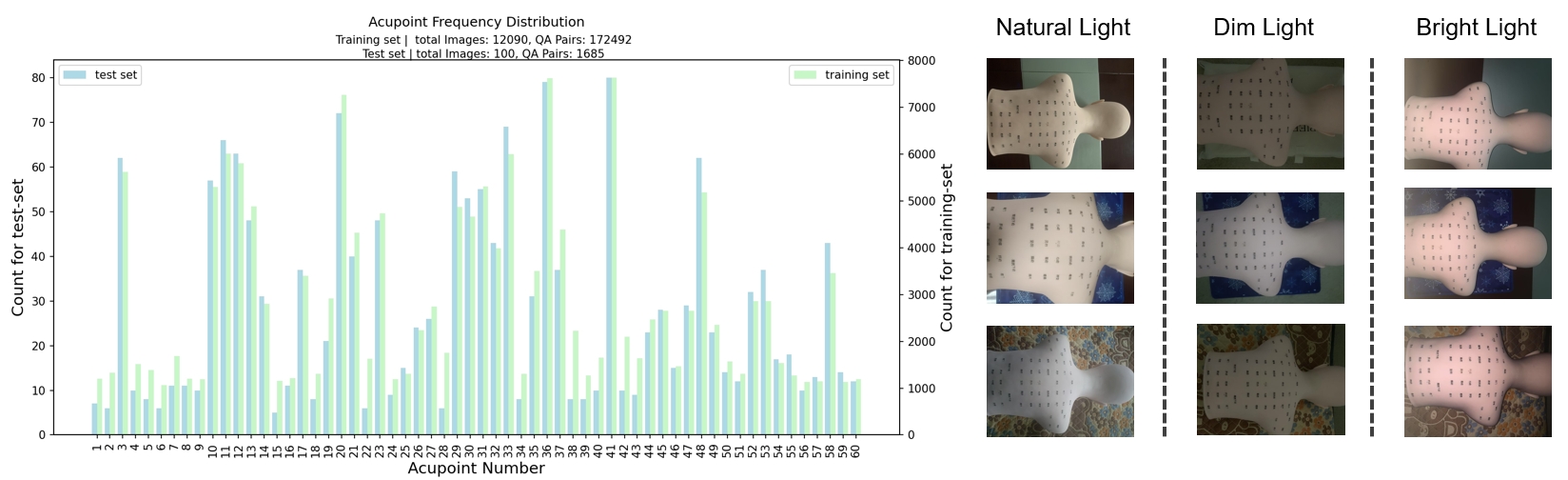}
  \caption{The bar chart on the left shows the distribution of acupoints in the test set and training set, while the right side displays images of the mannequin with acupoints under different lighting and background conditions.}
  \label{fig:dataset}
\end{figure*}
\begin{itemize}
    \item \textbf{Dataset Construction}: We construct MedMassage-12K, the first large-scale multimodal dataset for embodied massage, providing a foundation for training and evaluation.  
    \item \textbf{Framework Development}: We propose a hierarchical framework that bridges natural language understanding and robotic control, enabling precise acupoint localization and safe operations.  
    \item \textbf{Model Benchmarking and Fine-Tuning}: We evaluate existing MLLMs and establish a benchmark for embodied massage tasks, and fine-tune Qwen-VL to validate our approach.  
    \item \textbf{Physical Validation}: We conduct real-world experiments with a Franka Panda robot, demonstrating the framework’s reliability and generalizability.  
   \vspace{-0.23cm}
\end{itemize}
% \linenumbers

%%%%%%%%%%%%%%%%%%%%%%%%%%%%%%%%%%%%%%%%%%%%%%%%%%%%%%%%%%%%%%%%%%%%%%%%%%%%%%%
%%%%%%%%%%%%%%%%%%%%%%%%%%%%%%%%%%%%%%%%%%%%%%%%%%%%%%%%%%%%%%%%%%%%%%%%%%%%%%%

\section{Related Work}

\subsection{Embodied AI}

Traditional robotic motion control \cite{joshi2020robotic, andrychowicz2020learning} has primarily relied on state-based reinforcement learning for training. Subsequently, several studies have incorporated visual observations as inputs to enhance the understanding of complex environments \cite{zhang2024navid,xu2024mrftrans,xu2024deffusion}. For instance, Where2Act \cite{mo2021where2act} utilizes visual observations to predict actionable regions of objects, while AnyGrasp \cite{fang2023anygrasp} employs point cloud data for learning grasping poses. Inspired by the advancements in MLLMs, recent works have proposed vision-language-action models tailored for robotic manipulation. For example, ManipLLM \cite{li2024manipllm} introduces a three-level instruction fine-tuning framework to endow MLLMs with manipulation capabilities, and SC-MLLM  \cite{liu2023selfcorrected} incorporates a correction mechanism to improve performance. Additionally, RoboMamba \cite{liu2025robomamba} leverages the Mamba architecture to enhance model efficiency. Models such as RT-1 \cite{brohan2022rt1}, RT-2 \cite{brohan2023rt2}, and OpenVLA \cite{kim2024openvla} directly fine-tune large pretrained MLLMs to predict robot actions using the Open X-Embodiment dataset \cite{openx2023}. In contrast to these approaches, our work focuses on advancing the application of MLLMs in embodied massage scenarios, aiming to address the unique challenges and requirements of this domain.

\subsection{Multimodal Large Language Models}

Recent advancements in large language models (LLMs) \cite{touvron2023llama, floridi2020gpt3} have demonstrated remarkable capabilities in complex reasoning and human intention understanding. To further enhance their multimodal comprehension, several studies, such as LLaVA \cite{liu2023visual}, Qwen-VL \cite{wang2024qwen2vl}, and Videollama \cite{wang2024qwen2vl}, have begun to fine-tune these models using multimodal data for multimodal learning. These models have seen increasing adoption in various applications, including visual dialogue, scene understanding, and robotic task planning. Despite the considerable performance improvements in MLLMs, their embodied massage capabilities remain underexplored. Consequently, we propose a hierarchical embodied massage model designed to enhance the performance of MLLMs in embodied medical scenarios.

\subsection{Open-Source Datasets for Embodied Healthcare}
Open-source datasets play a pivotal role in advancing embodied healthcare technologies, offering critical resources for research and innovation despite their current limitations. Key datasets such as CaDIS\cite{grammatikopoulou2021cadis}, which provides annotated video images for cataract surgery and supports semantic segmentation for computer-assisted surgery systems, and MEDQA\cite{singhal2023large}, which offers a diverse range of medical exam questions to enhance natural language processing-based medical question-answering systems, have significantly contributed to the field. Additionally, multimodal datasets like MedTrinity-25M\cite{xie2024medtrinity}, with over 25 million richly annotated medical images, enable tasks such as image classification, visual question answering, and report generation, fostering the development of multimodal models for complex diagnostics and patient care. These datasets primarily concentrate on general medical image segmentation and medical question-answering tasks. However, for acupoint massage tasks, there is currently a lack of large-scale, open-source multimodal datasets. To bridge this gap, we propose a novel dataset focused on acupoint-related data, aiming to advance research in embodied massage tasks and contribute to the development of intelligent healthcare systems.
\section{Method}

\subsection{Task Definition}  
The embodied massage task requires a robotic system to understand high-level natural language instructions, precisely localize acupoints, and plan safe robotic motions. Formally, given a textual command $\mathcal{T}$ (e.g., \textit{``Locate the Zusanli acupoint and apply moderate pressure''}) with multimodal inputs $\big(I_{\text{RGB}} \in \mathbb{R}^{W \times H \times 3},\ I_{\text{Depth}} \in \mathbb{R}^{W \times H}\big)$, the system must generate a collision-free 6-DOF end-effector pose $\mathbf{p} = (x, y, z, \theta_x, \theta_y, \theta_z) \in \mathbb{R}^3 \times \mathbb{SO}(3)$ for the Franka Emika Panda robot. This pose integrates both positional coordinates $\mathbf{a}_{\text{pos}} \in \mathbb{R}^3$ and directional orientation $\mathbf{a}_{\text{dir}} \in \mathbb{R}^{3 \times 3}$. The task inherently bridges natural language understanding, visual perception, and robotic control for real-world human-robot collaboration.

\subsection{Data Collection and Augmentation}
% 我们介绍了 MedMassage-12K，这是首个用于分层按摩操作学习的大规模多模态数据集，包含 1,030 张医疗假人穴位图像和 15,092 个标注的问答对（QA pairs）。该数据集包含多样化背景和光照条件（如自然光、昏暗光、明亮光），确保覆盖广泛环境以促进泛化（如图 \ref{fig:dataset} 所示）。在穴位分布方面，数据集涵盖了 60 个不同的按摩点。图 \ref{fig:dataset} 展示了训练集和测试集中的分布情况。每个穴位在训练与测试中的频率不同，更好地反映了真实需求，保证了多样性与代表性。为方便起见，我们将穴位名称映射为数字索引，后文中用 ID 表示穴位名称。为了进一步提升多样性与鲁棒性，我们采用了数据增强，特别关注位置和高度的泛化。通过随机裁剪和旋转等几何变换，我们生成了额外训练数据，使图像数量增加至 12,190，QA 对增加至 174,177。这不仅提升了在不同光照与背景下的泛化能力，还增强了模型在位置与高度变化下的适应性，确保其能在多种真实场景中进行有效推理与任务执行。最终数据集的构成（如表 \ref{tab:dataset_split_v2} 所示）包含训练和测试的标注数据与增强数据。本数据集为评估用于医疗干预的具身模型提供了基础。
We introduce MedMassage-12K, the first large-scale multimodal dataset for hierarchical massage manipulation learning, which consists of 1,030 acupoint images of medical dummies and 15,092 annotated question and answer (QA) pairs. The dataset includes images from diverse backgrounds and lighting conditions (e.g., Natural, Dim, Bright), ensuring it covers a wide range of environments to promote generalization (as shown in Figure \ref{fig:dataset}). In terms of acupoint distribution, the dataset covers 60 different massage points. Figure \ref{fig:dataset} illustrates the distribution in training and test sets. The frequency of each acupoint varies between them, better reflecting real-world demand and ensuring diversity and representativeness. For convenience, we map acupoint names to numerical indices, and in the subsequent text, denote names by IDs. To enhance diversity and robustness, we applied data augmentation, focusing on generalization across position and height. Using geometric transformations such as random cropping and rotations, we generated additional training data. These augmentations enriched the dataset, increasing the images to 12,190 and QA pairs to 174,177. This not only improved generalization across lighting and backgrounds but also enhanced adaptability to positions and heights, ensuring effective reasoning in real-world scenarios. The final dataset composition, as shown in Table \ref{tab:dataset_split_v2}, includes both annotated and augmented data for training and testing. This dataset provides the foundation for evaluating embodied models for healthcare interventions.

\begin{table}[ht]
\centering
\caption{ MedMassage-12K Dataset Composition}
\vspace{0.25cm}
\label{tab:dataset_split_v2}
\begin{tabular}{@{}llrr@{}}
\toprule
\textbf{Category}  & \textbf{Partition} & \textbf{Images} & \textbf{QA Pairs} \\ 
\midrule
Training Set     &  Annotated Data        & 930             & 13,407            \\
                 &  Augmented Data           & 12,090             & 172,492             \\
\addlinespace
\midrule
   Test Set         &  Annotated Data       & 100         & 1,685           \\

\bottomrule
\end{tabular}
\vspace{-0.53cm}
\end{table}

\subsection{System Overview}

% 层级模型的总览
% 
% In this section, we demonstrate how we empower MLLMs
% with 穴位按摩。 The proposed model framework is illustrated in Figure2. 其中High-Level grounding Module 模块负责理解人类指令，并输出穴位位置。low-level control 模块，根据穴位位置转变为模型的控制量。
\begin{figure*}[t!]
  \centering
  \includegraphics[width=\textwidth]{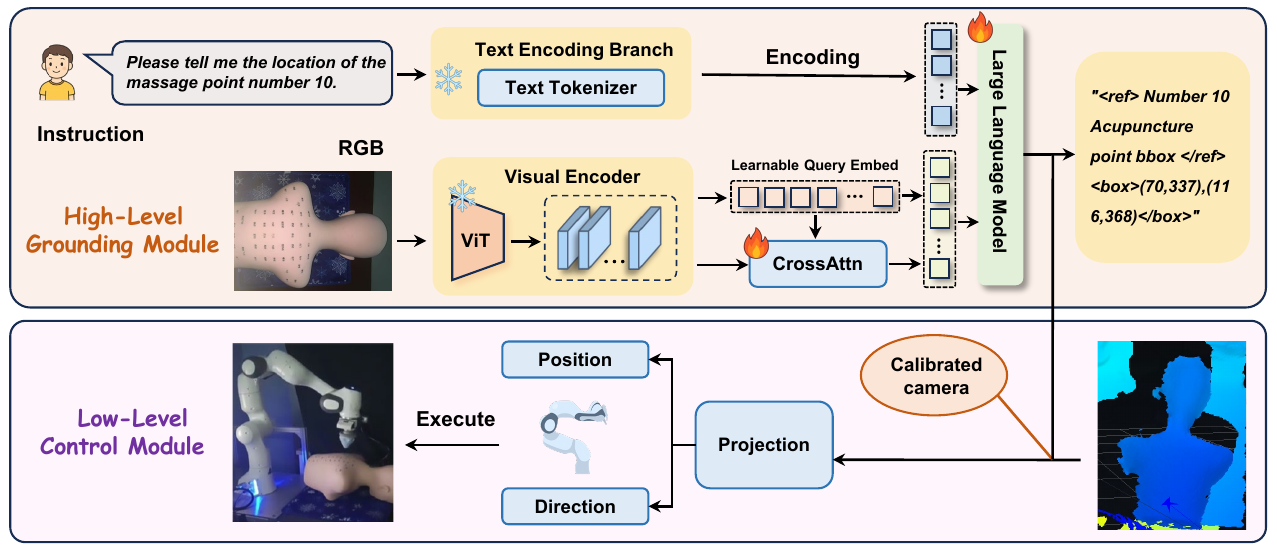}
  \caption{System architecture of the proposed framework, consisting of a High-Level Grounding Module (HLGM) and a Low-Level Control Module (LLCM).}
  \label{fig:model}
  \vspace{-0.53cm}
\end{figure*}

In this section, we demonstrate how we enable MLLMS to possess acupoint massage capabilities. As depicted in Fig \ref{fig:model}, our architecture consists of two core components: (1) a High-Level Grounding Module (HLGM) for semantic interpretation of human instructions and acupoint localization, and (2) a Low-Level Control Module (LLCM) that translates spatial coordinates into kinematic control signals.

\subsubsection{High-Level grounding Module}
% As for the high-level grounding model, we utilize the pretrained qwen vl. 
% 文本指令T通过预训练的Qwen 的tokennizer编码为文本特征。 
% 给定一个RGB 图像$$$I \in R^{H\times W \times 3}$ , 我们使用Openclip ViT-bigG 作为视觉提取器来提取视觉特征，
% 然后a single-layer cross-attention module as the vision-language adaptor, uses a goup of trainable vectors as qurey and the images fetures as keys for cross attention opretations.compresses the image features.

% after that, Qwen 7B is 作为LLM 解码器生成回答。 
For the high-level grounding model, we employ the pre-trained \texttt{Qwen-VL\cite{wang2024qwen2vl}}. The textual instruction \( T \) is encoded into textual features using the tokenizer of the pre-trained \texttt{Qwen} model. Given an RGB image \( I \in \mathbb{R}^{H \times W \times 3} \), we utilize OpenCLIP ViT-bigG \cite{radford2021learning} as the visual feature extractor to obtain visual representations. Subsequently, a single-layer cross-attention module serves as the vision-language adaptor. This module employs a set of trainable vectors as queries and the image features as keys for cross-attention operations, effectively compressing the image features.
Following this, the \texttt{Qwen-7B} model acts as the LLM decoder to generate the final response.  As illustrated in the Fig \ref{fig:model}, we freeze the visual encoder during training and only fine-tune the vision-language adaptor and the language model. The outputs of the model are supervised using the cross-entropy loss function.

During training, we utilize augmented data consisting of 172,492 QA pairs to train our model. Following the design of \texttt{Qwen-VL}, we use special tokens to distinguish between different types of information. For the input, we adopt the special token \texttt{<img>} to differentiate between image inputs and text inputs. For the output, to enhance the model’s fine-grained understanding and grounding capabilities for massage points, we use special tokens \texttt{<box>} and \texttt{</box>} to demarcate bounding box coordinates, and \texttt{<ref>} and \texttt{</ref>} to associate bounding boxes with their corresponding descriptive words or sentences. For any acupuncture point, the bounding box coordinates are normalized within the range \([0, 1000)\). These tokens ensure accurate localization and description of target regions, such as specific massage points in an image. The input and output format can be summarized as follows:

\begin{figure}[h]
\centering
  \includegraphics[width=0.5\textwidth]{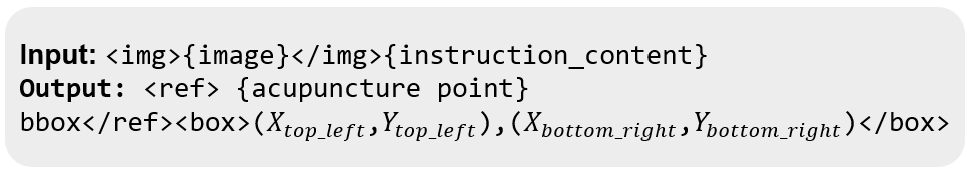}
  \vspace{-0.53cm}
\end{figure}

% \textbf{Input}:  
% \begin{verbatim}
% <img>{image path}</img>\nPlease tell me the location of the acupuncture point number 58.
% \end{verbatim}

% \textbf{Output}:  
% \begin{verbatim}
% <ref>Acupuncture point number 58 bbox</ref><box>(416,878),(459,901)</box>
% \end{verbatim}

\subsubsection{Low-Level control Module}
In this approach, the MLLM is first used to process the input 2D image and extract the 2D coordinates of the acupoint locations in the image. Then, a depth map of the scene is obtained using a depth camera, which provides the actual distance from the camera to each pixel. By utilizing the depth camera's intrinsic and extrinsic parameters, along with the depth information, the 2D pixel coordinates in the image can be transformed into corresponding 3D point cloud coordinates, namely position \( (x, y, z) \).These points can be used for plane fitting. We can apply the RANSAC algorithm to extract a plane model from the point cloud data, and subsequently obtain the normal vector of the plane. Next, we calculate the angle between the fitted plane's normal vector and the reference plane's normal vector using their dot product. Finally, by utilizing the "vertical tapping principle," we can deduce the object's rotational posture based on the change in the plane's normal vector, thereby obtaining the direction. Combining this with the previously obtained position \( (x, y, z) \), we can obtain the complete pose \( p = (x, y, z, \theta_x, \theta_y, \theta_z) \).

After obtaining the target's pose \( p = (x, y, z, \theta_x, \theta_y, \theta_z) \), the robotic arm calculates the corresponding joint configurations using Inverse Kinematics (IK) algorithms to move the end-effector to the target position while maintaining the correct orientation. This step ensures that the arm can accurately reach the target and perform precise operations. Next, the robotic arm uses the polynomial fitting to plan a collision-free path from the current joint configuration to the target position. Based on the path planning, the robotic arm generates a smooth trajectory using spline interpolation to ensure that the movement process does not involve abrupt acceleration or velocity changes. After trajectory generation, the robotic arm incorporates feedback control during execution to ensure it follows the planned trajectory. Finally, the robotic arm reaches the acupoint position and performs the massage operation.

\section{Experiments}
\subsection{Acupoint Grounding Success Rate Across Models} 

To evaluate the acupoint localization capabilities of existing MLLMs, we conducted comprehensive tests on Qwen-VL-Max and GPT-4o using the following standardized input instruction:
\begin{figure}[h!]
  \centering
  \includegraphics[width=0.5\textwidth]{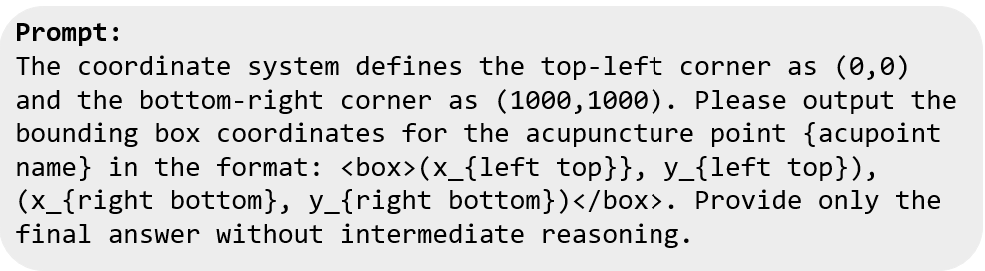}
  \label{fig:prompt}
   \vspace{-0.53cm}
\end{figure}

\begin{table}[h]
\centering

\caption{Acupoint grounding success across IoU thresholds.}
\vspace{0.25cm}
\begin{tabular}{|c|c|c|c|}
\hline
\multirow{2}{*}{Model} & \multicolumn{3}{c|}{Grounding Success Rate} \\ \cline{2-4} 
                      & IoU=0.3 & IoU=0.5 & IoU=0.75 \\ \hline
Qwen-VL-Max               & 0.12\%          & 0\%        &   0\%    \\ \hline
GPT-4o                  & 0.59\%           & 0.07\%        & 0\%       \\ \hline
Our Model                 & 87.60\%           & 81.42\%        & 67.77\%        \\ \hline
\end{tabular}
\label{tab:res1}
\vspace{-0.53cm}
\end{table}
As shown in Table \ref{tab:res1}, current multimodal large language models demonstrate significant limitations in acupoint grounding tasks. Both GPT-4o and Qwen-VL-Max exhibit grounding success rates approaching 0\% across various IoU thresholds, indicating their lack of fine-grained perception capability for massage acupoints. In contrast, our proposed model achieves substantially higher grounding success rates of 87.60\%, 81.42\%, and 67.77\% at IoU thresholds of 0.3, 0.5, and 0.75 respectively, demonstrating superior precision in acupoint localization tasks.

\subsection{Ablation study}

\subsubsection{Impact of Data Scale on Model Performance}  

In this section, we analyze the impact of data scale on model performance. We train the model using subsets of the annotated training data, specifically 10\%, 20\%, 40\%, 70\%, and 100\% of the original and augmented samples, and evaluate the performance on the same test set. 

\begin{table}[h!]
\caption{Impact of the Data Scale.}  
\vspace{0.15cm}
\centering
\begin{tabular}{|c|c|c|c|}
\hline
\multirow{2}{*}{Data} & \multicolumn{3}{c|}{Grounding Success Rate} \\ \cline{2-4} 
                      & IoU=0.3 & IoU=0.5 & IoU=0.75 \\ \hline
10\%                  & 47.18\%           & 37.45\%        & 26.17\%        \\ \hline
20\%                  & 61.60\%           & 51.10\%        & 39.17\%        \\ \hline
40\%                  & 75.25\%           & 66.83\%        & 53.41\%        \\ \hline
70\%                  & 83.03\%           & 75.91\%        & 62.49\%        \\ \hline
100\%                 & 87.60\%           & 81.42\%        & 67.77\%        \\ \hline
\end{tabular}
\label{tab:scale_res}
\vspace{-0.15cm}
\end{table}
Our experimental results demonstrate a clear trend: the model's performance consistently improves as the data scale increases, as shown in Table \ref{tab:scale_res}.
For instance, at an IoU threshold of 0.3, the test accuracy rises from 47.18\% with 10\% of the data to 87.60\% with the full dataset. Similar improvements are observed at higher IoU thresholds (0.5 and 0.75), indicating that larger datasets enhance the model’s ability to localize acupoints with greater precision.  
However, the incremental gains diminish as the data scale grows, particularly when moving from 70\% to 100\% of the dataset. This suggests that while increasing data volume is beneficial, the diversity and quality of the data also play a critical role in achieving further performance improvements. These findings highlight the importance of both data scale and diversity in developing robust models for acupoint localization tasks.  

\subsubsection{Impact of Data Augmentation}  

To evaluate the effectiveness of our data augmentation strategy, we conduct experiments with and without data augmentation. The results, as shown in Table \ref{table:aug}, demonstrate a significant improvement in grounding success rates when data augmentation is applied.  

\begin{table}[h!]
\centering
\caption{Grounding Success Rate w/ and w/o Data Aug.} 
\vspace{0.15cm}
\begin{tabular}{|c|c|c|c|}
\hline
\multirow{2}{*}{Data} & \multicolumn{3}{c|}{Grounding Success Rate} \\ \cline{2-4} 
                      & IoU=0.3 & IoU=0.5 & IoU=0.75 \\ \hline
w/o data augmentation  & 60.89\%           & 48.96\%        & 35.54\%        \\ \hline
W data augmentation  & 87.60\%           & 81.42\%        & 67.77\%        \\ \hline
\end{tabular}
\vspace{-0.3cm}
\label{table:aug}
\end{table}
Without data augmentation, the grounding success rates are 60.89\%, 48.96\%, and 35.54\% for IoU thresholds of 0.3, 0.5, and 0.75, respectively. With data augmentation, these rates increase significantly to 87.60\%, 81.42\%, and 67.77\% for the same thresholds. This represents performance improvements of 26.71 at IoU=0.3, 32.46 at IoU=0.5, and 32.23 at IoU=0.75. The substantial gains highlight the critical role of data augmentation in enhancing the model’s ability to generalize and accurately localize acupoints under diverse conditions.  

\subsection{Massage in Real-World Environments} 

\begin{figure}[h!]
  \centering
  \includegraphics[width=\linewidth]{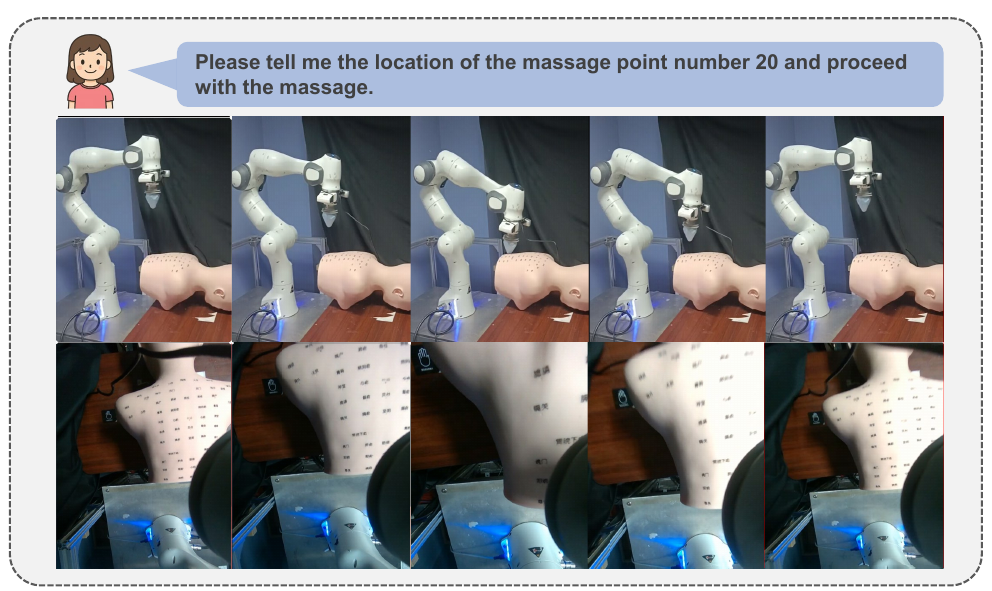}
  \caption{Massage in real-world environments: first-person RGB and third-person frames of robot execution.}
  \label{fig:result}
     \vspace{-0.13cm}
\end{figure}

To further evaluate the generalizability of our framework, we conduct experiments in real-world environments. We use a Franka Panda Robot equipped with a massage ball as the actuator and a RealSense D455 camera to capture RGB and depth images. The HMR model is deployed on a computer with an NVIDIA A6000 GPU.  As demonstrated in Fig \ref{fig:result}, our model can accurately locate and perform massage on specified acupuncture points under varying lighting conditions and backgrounds, showcasing the robustness and generalizability of our approach.

\section{Conclusions}

\label{sec:conclusions}
In this work, we address key challenges in embodied intelligence for healthcare by introducing a hierarchical framework and a large-scale multimodal dataset for acupoint massage tasks. Our framework integrates a high-level grounding module for acupoint localization and a low-level control module for trajectory planning. We evaluate and fine-tune multimodal large language models, demonstrating strong generalization across diverse conditions. Physical experiments validate the framework’s practical applicability, paving the way for robust and scalable healthcare solutions.

\bibliography{IF}

@article{singhal2023large,
  title={Large language models encode clinical knowledge},
  author={Singhal, Karan and Azizi, Shekoofeh and Tu, Tao and Mahdavi, S Sara and Wei, Jason and Chung, Hyung Won and Scales, Nathan and Tanwani, Ajay and Cole-Lewis, Heather and Pfohl, Stephen and others},
  journal={Nature},
  volume={620},
  number={7972},
  pages={172--180},
  year={2023},
  publisher={Nature Publishing Group}
}

@article{xie2024medtrinity,
  title={Medtrinity-25m: A large-scale multimodal dataset with multigranular annotations for medicine},
  author={Xie, Yunfei and Zhou, Ce and Gao, Lang and Wu, Juncheng and Li, Xianhang and Zhou, Hong-Yu and Liu, Sheng and Xing, Lei and Zou, James and Xie, Cihang and others},
  journal={arXiv preprint arXiv:2408.02900},
  year={2024}
}

@article{grammatikopoulou2021cadis,
  title={CaDIS: Cataract dataset for surgical RGB-image segmentation},
  author={Grammatikopoulou, Maria and Flouty, Evangello and Kadkhodamohammadi, Abdolrahim and Quellec, Gwenol{\'e} and Chow, Andre and Nehme, Jean and Luengo, Imanol and Stoyanov, Danail},
  journal={Medical Image Analysis},
  volume={71},
  pages={102053},
  year={2021},
  publisher={Elsevier}
}

@inproceedings{joshi2020robotic,
  title={Robotic grasping using deep reinforcement learning},
  author={Joshi, Shirin and Kumra, Sulabh and Sahin, Ferat},
  booktitle={2020 IEEE 16th International Conference on Automation Science and Engineering (CASE)},
  pages={1461--1466},
  year={2020},
  organization={IEEE}
}

@article{andrychowicz2020learning,
  title={Learning dexterous in-hand manipulation},
  author={Andrychowicz, Marcin and Baker, Bowen and Chociej, Maciek and Jozefowicz, Rafal and McGrew, Bob and Pachocki, Jakub and Petron, Arthur and Plappert, Matthias and Powell, Glenn and Ray, Alex and others},
  journal={The International Journal of Robotics Research},
  volume={39},
  number={1},
  pages={3--20},
  year={2020},
  publisher={SAGE Publications Sage UK: London, England}
}

@inproceedings{mo2021where2act,
  title={Where2act: From pixels to actions for articulated 3d objects},
  author={Mo, Kaichun and Guibas, Leonidas J and Mukadam, Mustafa and Gupta, Abhinav and Tulsiani, Shubham},
  booktitle={Proceedings of the IEEE/CVF International Conference on Computer Vision},
  pages={6813--6823},
  year={2021}
}

@article{fang2023anygrasp,
  title={Anygrasp: Robust and efficient grasp perception in spatial and temporal domains},
  author={Fang, Hao-Shu and Wang, Chenxi and Fang, Hongjie and Gou, Minghao and Liu, Jirong and Yan, Hengxu and Liu, Wenhai and Xie, Yichen and Lu, Cewu},
  journal={IEEE Transactions on Robotics},
  year={2023}
}

@inproceedings{li2024manipllm,
  title={Manipllm: Embodied multimodal large language model for object-centric robotic manipulation},
  author={Li, X and Zhang, M and Geng, Y and others},
  booktitle={Proceedings of the IEEE/CVF Conference on Computer Vision and Pattern Recognition},
  pages={18061--18070},
  year={2024}
}

@article{liu2025robomamba,
  title={RoboMamba: Efficient Vision-Language-Action Model for Robotic Reasoning and Manipulation},
  author={Liu, J and Liu, M and Wang, Z and others},
  journal={Advances in Neural Information Processing Systems},
  volume={37},
  pages={40085--40110},
  year={2025}
}

@article{liu2023selfcorrected,
  title={Self-Corrected Multimodal Large Language Model for Robot Manipulation and Reflection},
  author={Liu, J and Li, C and Wang, G and others},
  journal={Journal Name},
  year={2023}
}

@article{xu2024mrftrans,
  title={MRFTrans: Multimodal Representation Fusion Transformer for monocular 3D semantic scene completion},
  author={Xu, Rongtao and Zhang, Jiguang and Sun, Jiaxi and Wang, Changwei and Wu, Yifan and Xu, Shibiao and Meng, Weiliang and Zhang, Xiaopeng},
  journal={Information Fusion},
  pages={102493},
  year={2024},
  publisher={Elsevier}
}

@inproceedings{xu2024deffusion,
  title={DefFusion: Deformable Multimodal Representation Fusion for 3D Semantic Segmentation},
  author={Xu, Rongtao and Wang, Changwei and Zhang, Duzhen and Zhang, Man and Xu, Shibiao and Meng, Weiliang and Zhang, Xiaopeng},
  booktitle={2024 IEEE International Conference on Robotics and Automation (ICRA)},
  pages={7732--7739},
  year={2024},
  organization={IEEE}
}

@article{zhang2024navid,
  title={NaVid: Video-based VLM Plans the Next Step for Vision-and-Language Navigation},
  author={Zhang, Jiazhao and Wang, Kunyu and Xu, Rongtao and Zhou, Gengze and Hong, Yicong and Fang, Xiaomeng and Wu, Qi and Zhang, Zhizheng and He, Wang},
  journal={arXiv preprint arXiv:2402.15852},
  year={2024}
}

@inproceedings{openx2023,
  title={Open x-embodiment: Robotic learning datasets and rt-x models},
  author={Vuong, Quan and Levine, Sergey and Walke, Homer Rich and Pertsch, Karl and Singh, Anikait and Doshi, Ria and Xu, Charles and Luo, Jianlan and Tan, Liam and Shah, Dhruv and others},
  booktitle={Towards Generalist Robots: Learning Paradigms for Scalable Skill Acquisition@ CoRL2023},
  year={2023}
}

@article{brohan2023rt2,
  title={Rt-2: Vision-language-action models transfer web knowledge to robotic control},
  author={Brohan, Anthony and Brown, Noah and Carbajal, Justice and Chebotar, Yevgen and Chen, Xi and Choromanski, Krzysztof and Ding, Tianli and Driess, Danny and Dubey, Avinava and Finn, Chelsea and others},
  journal={arXiv preprint arXiv:2307.15818},
  year={2023}
}

@article{kim2024openvla,
  title={OpenVLA: An Open-Source Vision-Language-Action Model},
  author={Kim, {Moo Jin} and Pertsch, Karl and Karamcheti, Siddharth and others},
  journal={arXiv preprint arXiv:2406.09246},
  year={2024}
}

@article{brohan2022rt1,
  title={Rt-1: Robotics transformer for real-world control at scale},
  author={Brohan, A. and Brown, N. and Carbajal, J. and others},
  journal={arXiv preprint arXiv:2212.06817},
  year={2022}
}

@article{touvron2023llama,
  title={Llama: Open and efficient foundation language models},
  author={Touvron, Hugo and Lavril, Thibaut and Izacard, Gautier and Martinet, Xavier and Lachaux, Marie-Anne and Lacroix, Timothee and Roziere, Baptiste and Goyal, Naman and Hambro, Eric and Azhar, Faisal and others},
  journal={arXiv preprint arXiv:2302.13971},
  year={2023}
}

@article{floridi2020gpt3,
  title={GPT-3: Its nature, scope, limits, and consequences},
  author={Floridi, Luciano and Chiriatti, Massimo},
  journal={Minds and Machines},
  volume={30},
  pages={681--694},
  year={2020}
}

@article{liu2023visual,
  title={Visual instruction tuning},
  author={Liu, Haotian and Li, Chunyuan and Wu, Qingyang and Lee, Yong Jae},
  journal={arXiv preprint arXiv:2304.08485},
  year={2023}
}

@article{wang2024qwen2vl,
  title={Qwen2-vl: Enhancing vision-language model's perception of the world at any resolution},
  author={Wang, P. and Bai, S. and Tan, S. and others},
  journal={arXiv preprint arXiv:2409.12191},
  year={2024}
}

@inproceedings{radford2021learning,
  title={Learning transferable visual models from natural language supervision},
  author={Radford, Alec and Kim, Jong Wook and Hallacy, Chris and Ramesh, Aditya and Goh, Gabriel and Agarwal, Sandhini and Sastry, Girish and Askell, Amanda and Mishkin, Pamela and Clark, Jack and others},
  booktitle={International conference on machine learning},
  pages={8748--8763},
  year={2021},
  organization={PmLR}
}

@article{han2025multimodal,
title = {Multimodal fusion and vision–language models: A survey for robot vision},
author = {Xiaofeng Han and Shunpeng Chen and Zenghuang Fu and Zhe Feng and Lue Fan and Dong An and Changwei Wang and Li Guo and Weiliang Meng and Xiaopeng Zhang and Rongtao Xu and Shibiao Xu},
journal = {Information Fusion},
volume = {126},
pages = {103652},
year = {2026}
}

@article{ren2015faster,
  title={Faster r-cnn: Towards real-time object detection with region proposal networks},
  author={Ren, Shaoqing and He, Kaiming and Girshick, Ross and Sun, Jian},
  journal={Advances in neural information processing systems},
  volume={28},
  year={2015}
}

@article{yolov11,
  title={Yolov11: An overview of the key architectural enhancements},
  author={Khanam, Rahima and Hussain, Muhammad},
  journal={arXiv preprint arXiv:2410.17725},
  year={2024}
}

@inproceedings{RT-DETR,
  title={Detrs beat yolos on real-time object detection},
  author={Zhao, Yian and Lv, Wenyu and Xu, Shangliang and Wei, Jinman and Wang, Guanzhong and Dang, Qingqing and Liu, Yi and Chen, Jie},
  booktitle={Proceedings of the IEEE/CVF conference on computer vision and pattern recognition},
  pages={16965--16974},
  year={2024}
}

@article{jin2025efficient,
  title={Efficient multimodal large language models: A survey},
  author={Jin, Yizhang and Li, Jian and Gu, Tianjun and Liu, Yexin and Zhao, Bo and Lai, Jinxiang and Gan, Zhenye and Wang, Yabiao and Wang, Chengjie and Tan, Xin and others},
  journal={Visual Intelligence},
  volume={3},
  number={1},
  pages={27},
  year={2025},
  publisher={Springer}
}

@inproceedings{yang2025thinking,
  title={Thinking in space: How multimodal large language models see, remember, and recall spaces},
  author={Yang, Jihan and Yang, Shusheng and Gupta, Anjali W and Han, Rilyn and Fei-Fei, Li and Xie, Saining},
  booktitle={Proceedings of the Computer Vision and Pattern Recognition Conference},
  pages={10632--10643},
  year={2025}
}

@article{Rekep,
  title={Rekep: Spatio-temporal reasoning of relational keypoint constraints for robotic manipulation},
  author={Huang, Wenlong and Wang, Chen and Li, Yunzhu and Zhang, Ruohan and Fei-Fei, Li},
  journal={arXiv preprint arXiv:2409.01652},
  year={2024}
}

@article{ye2025deepseek,
  title={DeepSeek in healthcare: a survey of capabilities, risks, and clinical applications of open-source large language models},
  author={Ye, Jiancheng and Bronstein, Sophie and Hai, Jiarui and Hashish, Malak Abu},
  journal={arXiv preprint arXiv:2506.01257},
  year={2025}
}

@article{longhini2025unfolding,
  title={Unfolding the literature: A review of robotic cloth manipulation},
  author={Longhini, Alberta and Wang, Yufei and Garcia-Camacho, Irene and Blanco-Mulero, David and Moletta, Marco and Welle, Michael and Aleny{\`a}, Guillem and Yin, Hang and Erickson, Zackory and Held, David and others},
  journal={Annual Review of Control, Robotics, and Autonomous Systems},
  volume={8},
  number={1},
  pages={295--322},
  year={2025},
  publisher={Annual Reviews}
}

@misc{chapman2018evidence,
  title={Evidence-based medicine, media, and manipulation},
  author={Chapman, Jens R and Wiechert, Karsten and Wang, Jeffrey C},
  journal={Global Spine Journal},
  volume={8},
  number={5},
  pages={437--439},
  year={2018},
  publisher={SAGE Publications Sage CA: Los Angeles, CA}
}

@article{Rdt-1b,
  title={Rdt-1b: a diffusion foundation model for bimanual manipulation},
  author={Liu, Songming and Wu, Lingxuan and Li, Bangguo and Tan, Hengkai and Chen, Huayu and Wang, Zhengyi and Xu, Ke and Su, Hang and Zhu, Jun},
  journal={arXiv preprint arXiv:2410.07864},
  year={2024}
}

@article{cleary2019manipulation,
  title={Manipulation in health care: A positive or negative experience?},
  author={Cleary, Michelle and West, Sancia and McGarry, Denise and Greenwood, Melanie and Kornhaber, Rachel},
  journal={Issues in Mental Health Nursing},
  volume={40},
  number={11},
  pages={985--987},
  year={2019},
  publisher={Taylor \& Francis}
}

@article{black2410pi0,
  title={$\pi$0: A vision-language-action flow model for general robot control. CoRR, abs/2410.24164, 2024. doi: 10.48550},
  author={Black, Kevin and Brown, Noah and Driess, Danny and Esmail, Adnan and Equi, Michael and Finn, Chelsea and Fusai, Niccolo and Groom, Lachy and Hausman, Karol and Ichter, Brian and others},
  journal={arXiv preprint ARXIV.2410.24164}
}

@article{hou2025dita,
  title={Dita: Scaling diffusion transformer for generalist vision-language-action policy},
  author={Hou, Zhi and Zhang, Tianyi and Xiong, Yuwen and Duan, Haonan and Pu, Hengjun and Tong, Ronglei and Zhao, Chengyang and Zhu, Xizhou and Qiao, Yu and Dai, Jifeng and others},
  journal={arXiv preprint arXiv:2503.19757},
  year={2025}
}

@article{hurst2024gpt,
  title={Gpt-4o system card},
  author={Hurst, Aaron and Lerer, Adam and Goucher, Adam P and Perelman, Adam and Ramesh, Aditya and Clark, Aidan and Ostrow, AJ and Welihinda, Akila and Hayes, Alan and Radford, Alec and others},
  journal={arXiv preprint arXiv:2410.21276},
  year={2024}
}

@article{ma2025phyblock,
  title={Phyblock: A progressive benchmark for physical understanding and planning via 3d block assembly},
  author={Ma, Liang and Wen, Jiajun and Lin, Min and Xu, Rongtao and Liang, Xiwen and Lin, Bingqian and Ma, Jun and Wang, Yongxin and Wei, Ziming and Lin, Haokun and others},
  journal={arXiv preprint arXiv:2506.08708},
  year={2025}
}

@inproceedings{xu2025a0,
  title={A0: An affordance-aware hierarchical model for general robotic manipulation},
  author={Xu, Rongtao and Zhang, Jian and Guo, Minghao and Wen, Youpeng and Yang, Haoting and Lin, Min and Huang, Jianzheng and Li, Zhe and Zhang, Kaidong and Wang, Liqiong and others},
  booktitle={Proceedings of the IEEE/CVF International Conference on Computer Vision},
  pages={13491--13501},
  year={2025}
}

@article{lin2023advances,
  title={Advances in embodied navigation using large language models: A survey},
  author={Lin, Jinzhou and Gao, Han and Feng, Xuxiang and Xu, Rongtao and Wang, Changwei and Zhang, Man and Guo, Li and Xu, Shibiao},
  journal={arXiv preprint arXiv:2311.00530},
  year={2023}
}

@article{lin2025evolvenav,
  title={EvolveNav: Self-Improving Embodied Reasoning for LLM-Based Vision-Language Navigation},
  author={Lin, Bingqian and Nie, Yunshuang and Zai, Khun Loun and Wei, Ziming and Han, Mingfei and Xu, Rongtao and Niu, Minzhe and Han, Jianhua and Lin, Liang and Lu, Cewu and others},
  journal={arXiv preprint arXiv:2506.01551},
  year={2025}
}

@article{zhang2025activevln,
  title={ActiveVLN: Towards Active Exploration via Multi-Turn RL in Vision-and-Language Navigation},
  author={Zhang, Zekai and Zhu, Weiye and Pan, Hewei and Wang, Xiangchen and Xu, Rongtao and Sun, Xing and Zheng, Feng},
  journal={arXiv preprint arXiv:2509.12618},
  year={2025}
}

@article{zhang2025robridge,
  title={RoBridge: A Hierarchical Architecture Bridging Cognition and Execution for General Robotic Manipulation},
  author={Zhang, Kaidong and Xu, Rongtao and Ren, Pengzhen and Lin, Junfan and Wu, Hefeng and Lin, Liang and Liang, Xiaodan},
  journal={arXiv preprint arXiv:2505.01709},
  year={2025}
}

@article{xu20253d,
  title={3D-MoRe: Unified Modal-Contextual Reasoning for Embodied Question Answering},
  author={Xu, Rongtao and Gao, Han and Yu, Mingming and An, Dong and Chen, Shunpeng and Wang, Changwei and Guo, Li and Liang, Xiaodan and Xu, Shibiao},
  journal={arXiv preprint arXiv:2507.12026},
  year={2025}
}

@article{yan2024instrugen,
  title={InstruGen: Automatic Instruction Generation for Vision-and-Language Navigation Via Large Multimodal Models},
  author={Yan, Yu and Xu, Rongtao and Zhang, Jiazhao and Li, Peiyang and Liang, Xiaodan and Yin, Jianqin},
  journal={arXiv preprint arXiv:2411.11394},
  year={2024}
}
\end{sloppypar}
\end{document}